\documentclass[conference,letterpaper]{IEEEtran}
\IEEEoverridecommandlockouts
\usepackage[top=0.8in, bottom=1.05in, left=0.65in, right=0.65in]{geometry}

\usepackage{subcaption}
\usepackage{graphicx}
\usepackage{ulem}
\usepackage{cite}
\usepackage{amsmath,amssymb,amsfonts}
\usepackage{algorithmic}
\usepackage{graphicx}
\usepackage{textcomp}
\usepackage{xcolor}
\def\BibTeX{{\rm B\kern-.05em{\sc i\kern-.025em b}\kern-.08em
    T\kern-.1667em\lower.7ex\hbox{E}\kern-.125emX}}

\usepackage{algorithm}
\usepackage{multirow}
\usepackage{bbding}
\usepackage{array}
\usepackage{makecell}
\usepackage{amsmath}
\usepackage{xspace}
\usepackage{color}
\usepackage{amssymb}
\usepackage{booktabs}
\usepackage{pifont}
\usepackage{bm}
\useunder{\uline}{\ul}{}

\usepackage{amsmath}    
\usepackage{xcolor}     

\usepackage{newfloat}
\usepackage{listings}
\def\eg{\textit{e.g.}\xspace}

\def\name{\textbf{GLSDA}\xspace}

\begin{document}

\title{
Generalizing WiFi Gesture Recognition \\via Large-Model-Aware Semantic \\Distillation and Alignment
\thanks{* Both authors contributed equally to this research.}
\thanks{† Corresponding Author. E-mail: hjy@hfut.edu.cn}

}

\author{
\IEEEauthorblockN{Feng-Qi Cui\textsuperscript{1,2*}, Yu-Tong Guo\textsuperscript{2*}, Tianyue Zheng\textsuperscript{3}, Jinyang Huang\textsuperscript{2†} }

\IEEEauthorblockA{\textit{\textsuperscript{1}Institute of Advanced Technology, University of Science and Technology of China,}
Hefei, China \\
\textit{\textsuperscript{2}School of Computer Science and Information Engineering, Hefei University of Technology,}
Hefei, China \\
\textit{\textsuperscript{3}School of Computer Science and Engineering, Southern University of Science and Technology,}
Shenzhen, China \\
Emails: fengqi\_cui@mail.ustc.edu.cn, hfutgyt@163.com, \\zhengty@sustech.edu.cn, hjy@hfut.edu.cn }
}

\maketitle

\begin{abstract}
WiFi-based gesture recognition has emerged as a promising RF sensing paradigm for enabling non-contact and privacy-preserving human-computer interaction in AIoT environments. However, existing methods often suffer from limited generalization and semantic expressiveness due to the domain-sensitive nature of Channel State Information and the lack of high-level gesture abstraction. To address these challenges, we propose a novel generalization framework, termed Large-Model-Aware Semantic Distillation and Alignment (\name), which leverages the semantic prior of pre-trained large foundation models to enhance gesture representation learning in both in-domain and cross-domain scenarios.
Specifically, we first design a dual-path CSI encoding pipeline that captures geometric and dynamic gesture patterns via CSI-Ratio phase sequences and Doppler spectrograms. These representations are then fed into a Multiscale Semantic Encoder, which learns robust temporal embeddings and aligns them with gesture semantics through cross-modal attention mechanisms. To further enhance category discrimination, we introduce a Semantic-Aware Soft Supervision scheme that encodes inter-class correlations and reduces label ambiguity, especially for semantically similar gestures. Finally, we develop a Robust Dual-Distillation strategy to compress the aligned model into a lightweight student network, jointly distilling intermediate features and semantic-informed soft labels from the teacher model.
Extensive experiments on the Widar3.0 benchmark show that \name~consistently outperforms state-of-the-art methods in both in-domain and cross-domain gesture recognition tasks, while significantly reducing model size and inference latency. Our method offers a scalable and deployable solution for generalized RF-based gesture interfaces in real-world AIoT applications.
\end{abstract}

\begin{IEEEkeywords}
Gesture recognition, channel state information, pre-trained large model, knowledge distillation, multimodal alignment.
\end{IEEEkeywords}

\section{Introduction}

WiFi-based gesture recognition has gained increasing attention as a non-intrusive and privacy-preserving sensing modality for human-computer interaction, especially in AIoT environments such as smart homes, ambient interfaces, and ubiquitous computing~\cite{cui2025learningheterogeneitygeneralizingdynamic,zhang2018widar3,10090421,10804189}. Unlike traditional vision-based~\cite{molchanov2015hand} or wearable-device-based methods, RF-based sensing enables contact-free interaction without requiring line-of-sight or additional hardware, making it particularly attractive for low-cost, passive, and context-aware applications~\cite{9613773,10899398,10508334,10587029}.

Recent advances have shown that Channel State Information (CSI) extracted from commercial WiFi signals can be effectively used to model human gestures by capturing the fine-grained temporal and spatial perturbations induced by hand or body movements~\cite{10149418,10090421,feng2025rf}. However, despite achieving competitive accuracy in constrained scenarios, existing WiFi gesture recognition methods often suffer from severe performance drops when deployed across spatial domains with varying user locations, body orientations, and ambient environments~\cite{wang2022airfi,yang2022autofi,10916982,10339891}. This is mainly due to the inherent domain sensitivity of CSI signals and the lack of generalizable representation learning mechanisms~\cite{10804189,10699348,10916982}.

Despite recent progress in WiFi-based gesture recognition, achieving robust generalization across varying spatial domains remains a significant challenge. We identify three fundamental obstacles that hinder the deployment of such systems in real-world AIoT scenarios:  
\uline{1) Domain-sensitive CSI dynamics degrade signal consistency.} The temporal and spatial fluctuations of Channel State Information (CSI) are highly susceptible to changes in antenna placement, user orientation, and environmental reflectors \cite{9613773,9075376,10804189}. These variations introduce domain-specific distortions in both phase and amplitude, which dramatically shift gesture signal patterns across locations or sessions. As a result, existing models trained on fixed environments often overfit to domain-specific features and fail to generalize to unseen setups.  
\uline{2) Lightweight models lack the capacity for semantic abstraction.} To enable deployment on resource-constrained AIoT devices, gesture recognition models must be compact and efficient \cite{han2015deepcompression,fan2024multi,10251628,10508334}. However, smaller models inherently struggle to capture high-level semantics or encode domain-invariant representations. This limitation becomes especially severe in ambiguous gesture classes or under cross-domain shifts, where simple pattern matching fails to capture semantic equivalence~\cite{guo2024benchmarking,hu2025unified}.  
\uline{3) Discrete labels provide insufficient semantic supervision.} Conventional training pipelines rely on one-hot gesture labels that treat all classes as equally distant, ignoring inherent semantic correlations (\eg, 'push' vs. 'press'). Without such semantic awareness, models lack the necessary guidance to align related gestures across domains, leading to representational drift and reduced recognition robustness in noisy or low-resource conditions~\cite{guo2021context,zhou2022contrastive}.  
These challenges collectively suggest the need for a general framework that learns robust and domain-invariant gesture features, incorporates semantic knowledge to guide representation learning, and distills such knowledge into deployable lightweight models.

To address the above challenges in WiFi-based gesture recognition, inspired by the principles of knowledge distillation~\cite{hinton2015distilling} and modality alignment~\cite{tsai2019multimodal}, we propose a novel framework named Generalizing WiFi Gesture Recognition via Large-Model-Aware Semantic Distillation and Alignment (\name), which incorporates two plug-and-play gesture-enhancing modules that can be seamlessly integrated into existing WiFi recognition backbones. 
First, to effectively encode semantic priors from vision-language knowledge and improve WiFi representation generalization, we design the Large-Model Semantic Distillation Module (LSDM). By leveraging weakly-paired supervision from synchronized video and WiFi data, LSDM extracts conceptual gesture semantics from large-scale pre-trained foundation models. These high-level cues are then distilled into the WiFi feature space to provide meaningful semantic guidance, enabling structure-aware gesture encoding and semantic preservation across environments.
To further reduce domain discrepancy and improve the temporal robustness of WiFi representations, we introduce the Modality-Aligned Representation Optimization Module (MARO). By performing temporal-aware distribution alignment and representation smoothing in both feature and prediction spaces, MARO encourages domain-invariant feature learning while preserving discriminative capacity. This dual-view optimization enhances cross-domain generalization and representation stability, especially in unseen subject and location scenarios.
Apparently, by synergizing the semantic abstraction capability of large models with tailored WiFi-domain alignment mechanisms, \name substantially improves gesture recognition robustness without requiring any additional labeled data from the corresponding target domain.

Totally, our contributions can be summarized as follows:
\begin{itemize}
\item We present \name, the first generalization framework that leverages large pre-trained foundation models (\eg, LLMs or VLMs) to guide WiFi-based gesture recognition. By distilling semantic priors and aligning modality-specific representations, our method bridges the semantic gap between discrete gesture labels and noisy CSI inputs.
\item We propose the Large-Model Semantic Distillation Module (LSDM) that transfers gesture-level conceptual knowledge from pretrained foundation models to the RF domain using weak vision-RF pairing. This enhances the semantic grounding and class-level consistency of WiFi representations.
\item We design the Modality-Aligned Representation Optimization Module (MARO) that performs cross-modal and cross-domain adaptation via distribution alignment and regularization, ensuring temporal consistency and spatial generalization of gesture features.
\item We conduct extensive experiments on the public WiGest and UbiGesture benchmarks, demonstrating that \name consistently outperforms prior methods in both cross-subject and cross-location settings, achieving superior generalization and inference efficiency.
\end{itemize}

\section{Related Work}

\subsection{WiFi-based Gesture Recognition and Various Domain Generalization}

WiFi-based gesture recognition has emerged as a promising paradigm for privacy-preserving and device-free human-computer interaction, by leveraging Channel State Information (CSI) to capture subtle human body dynamics. Early efforts focused on handcrafted feature extraction (\eg, amplitude, phase, spectrogram), followed by shallow classifiers~\cite{9613773}. With the rise of deep learning, a surge of learning-based approaches demonstrated superior recognition accuracy under constrained settings~\cite{yang2023sensefi,10090421}. Recent work has also explored multimodal extensions, combining WiFi sensing with vision for joint emotion or gesture recognition~\cite{10149418}, or even targeting respiratory and health-related monitoring tasks~\cite{10699348,10804189}. 

Nevertheless, CSI features are highly sensitive to environmental and domain shifts, causing significant performance degradation in unseen settings. To mitigate this, a variety of domain generalization methods have been proposed. AutoFi~\cite{yang2022autofi} leverages geometric self-supervised learning to learn invariant CSI embeddings across rooms. AirFi~\cite{wang2022airfi} aggregates environment-agnostic features without requiring target domain data. WiOpen~\cite{10899398} further addresses open-set recognition in WiFi-based gesture recognition by reducing uncertainty under domain shift. Other anti-interference works, such as PhaseAnti~\cite{9613773} and Subcarrier Correlation Selection~\cite{9075376} improve robustness against noise and co-channel interference. However, these approaches mainly emphasize low-level signal invariance and task-specific supervision. They lack semantic-level abstraction, which is critical for capturing conceptual similarity (\eg, “push” vs. “press”) across environments.

Our work complements this line by bridging RF gesture recognition with semantic priors from large-scale pretrained models, aiming to provide robust and interpretable generalization across domains.

\subsection{Knowledge Distillation and Large Pre-trained Foundation Model Semantic Transfer}

Knowledge Distillation (KD) has become a popular strategy to compress large models into efficient student models while retaining performance~\cite{sanh2019distilbert}. Beyond matching logits, recent studies highlight structured or multi-aspect semantic transfer. For example, multi-aspect KD enables transferring richer conceptual features, while distillation objectives tailored for large autoregressive LLMs (\eg, DistiLLM) improve fidelity in student models~\cite{ko2024distillm,lee2025quantification}. In the vision and multimodal domains, contrastive distillation and cross-modal semantic alignment have been widely explored~\cite{zhou2022contrastive,guo2021context,hu2025unified}. These methods show that large foundation models such as CLIP or BLIP can provide strong semantic anchors for smaller models to ground their representations.
On the other hand, research in model compression and efficiency also contributes to enabling KD in resource-constrained environments. For instance, multi-objective convex quantization~\cite{fan2024multi} has been proposed for efficient model deployment. In video understanding, distillation has been employed for action and micro-action recognition tasks~\cite{guo2024benchmarking}, demonstrating the effectiveness and reliability of high-level semantic transfer in temporal modeling tasks.

Despite these advances, the application of KD to RF sensing remains largely unexplored. Recent RF studies, including WiFi-based open-set recognition~\cite{10899398}, side-channel privacy threats~\cite{10587029}, and cross-modal sensing applications~\cite{10149418, feng2025rf}, show the need for both robust generalization and semantic interpretability. Yet, no prior work has investigated semantic distillation from large models into the RF domain. Thus, it is very meaningful to improve the generalization of WiFi sensing by introducing large models. 

\subsection{Positioning of Our Work}

In summary, prior works in WiFi-based gesture recognition have advanced CSI modeling, domain generalization, and anti-interference robustness, but largely neglect semantic alignment across modalities. On the other hand, knowledge distillation from large models has proven effective in NLP, vision, and multimodal understanding~\cite{guo2021context,zhou2022contrastive,hu2025unified}, but its potential in RF sensing remains untapped. Our proposed framework, \name, introduces a novel semantic distillation pipeline where large-scale pre-trained foundation models act as teachers to guide lightweight RF encoders via weak vision-RF pairing and modality-aligned optimization. This design enables cross-domain generalization, semantic abstraction, and deployment efficiency, addressing the key limitations of existing methods.

\begin{figure}[t]
\centering
\includegraphics[width=0.48\textwidth]{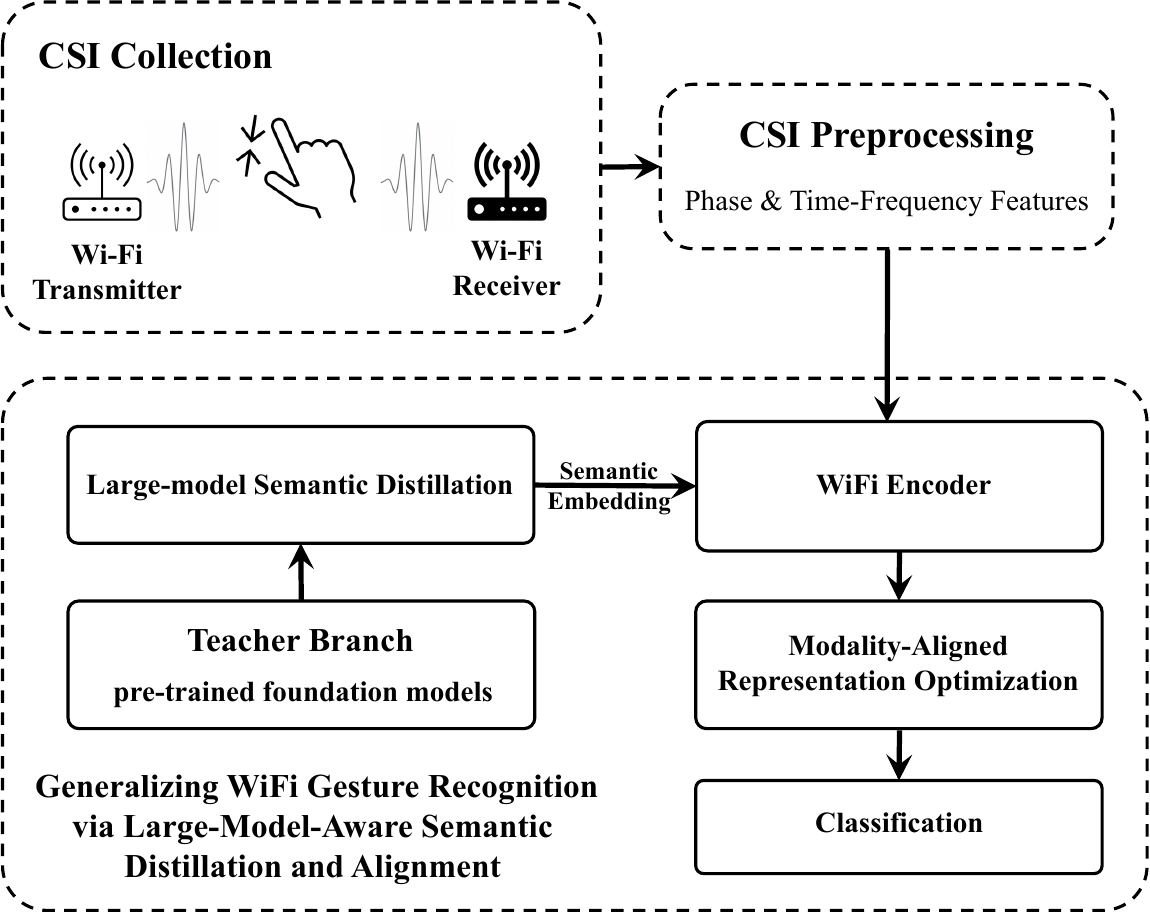} 
\vspace{0.08in}
\caption{Overview of the proposed \name framework for WiFi-based gesture recognition. The system takes CSI sequences as input, which are preprocessed into complementary time-frequency and phase features. A pre-trained foundation model branch provides semantic embeddings from synchronized visual data, serving as a teacher for the Large-model Semantic Distillation Module (LSDM). The distilled semantic priors guide the WiFi encoder to learn semantically meaningful representations. Meanwhile, the Modality-Aligned Representation Optimization Module (MARO) further reduces cross-modal discrepancy and temporal inconsistency through feature alignment and classifier smoothing. Finally, the optimized WiFi representations are fed into a classifier for robust gesture recognition.}
\label{ours}
\end{figure}

\section{Method}
To enhance the robustness and generalizability of WiFi-based gesture recognition across diverse environments and subjects, we propose a novel framework named \name. As illustrated in Fig.~\ref{ours}, the framework comprises two synergistic modules: a Large-model Semantic Distillation Module (LSDM) and a Modality-Aligned Representation Optimization Module (MARO).
Given synchronized pairs of WiFi CSI sequences and video frames, \name leverages  large pretrained models to extract rich semantic embeddings from large-scale foundation models. These high-level representations serve as soft targets to supervise the WiFi encoder via semantic distillation in LSDM. This enables the WiFi branch to inherit abstract and robust semantic priors from large-scale pre-trained foundation models  without requiring paired textual annotations.
Meanwhile, MARO is designed to address the modality discrepancy and distributional shifts inherent in cross-modal learning. It introduces a representation-level and classifier-level alignment mechanism to encourage consistent and discriminative gesture representations in the WiFi domain.
Through joint optimization of both modules, \name effectively bridges the semantic gap between WiFi and vision modalities, while promoting discriminative and transferable features under weakly paired supervision.

\subsection{Transferring Large Pre-trained Foundation Model Semantic Priors into WiFi Domain}

To bridge the semantic gap between WiFi signals and human-understandable concepts, we propose the LSDM, which introduces large model (LM) guided supervision to enhance the semantic representation capability of WiFi-based gesture features. This module enables semantic knowledge transfer by aligning feature embeddings from WiFi CSI and corresponding visual inputs under weak supervision. LSDM operates during training only and does not require visual input at inference time, thereby effectively saving time.

\subsubsection{Large Model Guided Semantic Supervision}  
To achieve high-level semantic alignment, we leverage gesture category labels as textual prompts and project them into the semantic space of a pretrained large model. Specifically, each gesture label is first converted into a natural language description (\eg, `a person performing a push gesture'), which is then fed into the text encoder $f_{\text{LM}}(\cdot)$ of a foundation model to obtain a semantic embedding:
\begin{equation}
\mathbf{z}_{\text{LM}} = f_{\text{LM}}(\text{Prompt}) \in \mathbb{R}^{d},
\end{equation}
where $\mathbf{z}_{\text{LM}}$ encodes rich conceptual priors distilled from large-scale language pretraining. These embeddings act as semantic anchors to guide the WiFi encoder, which effectively encourages it to capture domain-invariant and semantically meaningful gesture representations.

Simultaneously, the WiFi CSI signal $\mathbf{X}_{\text{CSI}} \in \mathbb{R}^{T \times C}$ is processed by a WiFi encoder $f_\theta(\cdot)$ to produce a corresponding feature embedding:
\begin{equation}
\mathbf{z}_{\text{CSI}} = f_\theta(\mathbf{X}_{\text{CSI}}) \in \mathbb{R}^{d},
\end{equation}
which is projected into the same semantic space as $\mathbf{z}_{\text{LM}}$.

\subsubsection{Contrastive Semantic Distillation.}  
To transfer semantic knowledge from LM to CSI features, we employ a contrastive loss to minimize the distance between $\mathbf{z}_{\text{CSI}}$ and $\mathbf{z}_{\text{LM}}$, while pushing apart mismatched pairs in the batch. We adopt a normalized temperature-scaled cross-entropy loss, which can be expressed as:
\begin{equation}
\mathcal{L}_{\text{LSDM}} = -\log \frac{\exp(\text{sim}(\mathbf{z}_{\text{CSI}}, \mathbf{z}_{\text{LM}})/\tau)}{\sum\limits_{j=1}^{N} \exp(\text{sim}(\mathbf{z}_{\text{CSI}}, \mathbf{z}_{\text{LM}}^j)/\tau)},
\end{equation}
where $\text{sim}(\cdot, \cdot)$ denotes cosine similarity, $\tau$ is a temperature scaling parameter, and $N$ is the number of negative pairs in the batch.
This formulation allows the WiFi encoder to absorb multimodal semantics through feature-level distillation, enabling improved generalization under cross-domain and cross-user settings. Notably, this process requires only synchronized CSI-visual pairs, and no gesture category labels are needed.

\subsection{ Reducing Cross-modal and Cross-domain Discrepancy}

While LSDM injects semantic priors from pre-trained foundation models into the WiFi domain, distributional discrepancies between modalities and environments still hinder the robustness of CSI-based representations. To further enhance generalization, we introduce the Modality-Aligned Representation Optimization Module (MARO), which explicitly enforces alignment in both feature and classifier spaces. MARO ensures that WiFi features not only capture semantically meaningful information but also remain domain-invariant under environmental shifts.

\begin{table*}[htbp]
\renewcommand{\arraystretch}{1.7}
\centering
\small
\setlength{\tabcolsep}{8mm}
\caption{Performance comparison of different methods on the Widar3.0 dataset. The table shows the accuracy under three scenarios: In-Domain (ID), Cross-Location (CL), and Cross-Orientation (CO). (\textbf{Bold}: Best result, \underline{Underline}: Second-best result)}
\label{tab:widar-results}
\scalebox{1.0}{
\begin{tabular}{c|ccc|c}
\hline
\multirow{2}{*}{\textbf{Methods}} & \multicolumn{3}{c|}{\textbf{Accuracy}} & \multirow{2}{*}{\textbf{Mean}} \\ 
\cmidrule(lr){2-4}
& \textbf{ID} & \textbf{CL} & \textbf{CO} & \\ 
\hline
CNN+GRU\cite{zhang2018widar3} & 92.7\% & 90.48\% & 81.58\% & 88.25\%\\
WiSR\cite{liu2023wisr} & - & 67.73\% & 69.74\% & 68.74\%\\
WiHF\cite{li2020wihf} & 97.65\% & 91.22\% & 80.64\% & 89.84\%\\
WiGNN\cite{chen2024wiggn} & - & \underline{95.20\%} & \textbf{93.30\%} & \underline{94.25\%}\\
THAT\cite{li2021twostream} & - & 71.56\% & 81.76\% & 76.66\%\\
\name(Ours) & \textbf{97.78\%} & \textbf{95.59\%} & \underline{92.80\%} & \textbf{95.39\%}\\
\hline
\end{tabular}
}
\end{table*}

\subsubsection{Feature-level Distribution Alignment}  
Let $\mathbf{z}_{\text{CSI}} \in \mathbb{R}^{d}$ denote the feature embedding from the WiFi encoder, and $\mathbf{z}_{\text{LM}} \in \mathbb{R}^{d}$ the corresponding semantic embedding distilled from the large model teacher. To minimize distribution mismatch between the two modalities, we introduce a discrepancy regularization loss:
\begin{equation}
\mathcal{L}_{\text{feat}} = \left\| \frac{1}{B}\sum_{i=1}^B \mathbf{z}_{\text{CSI}}^{(i)} - \frac{1}{B}\sum_{i=1}^B \mathbf{z}_{\text{LM}}^{(i)} \right\|_2^2,
\end{equation}
where $B$ is the batch size. This alignment encourages WiFi embeddings to preserve the semantic center of the teacher semantic space, thereby reducing modality-induced drift. 

To further stabilize temporal CSI dynamics, we apply a consistency regularization term between neighboring segments $\mathbf{z}_{\text{CSI}}^{t}$ and $\mathbf{z}_{\text{CSI}}^{t+1}$, formulated as:
\begin{equation}
\mathcal{L}_{\text{temp}} = \frac{1}{T-1}\sum_{t=1}^{T-1} \|\mathbf{z}_{\text{CSI}}^{t} - \mathbf{z}_{\text{CSI}}^{t+1}\|_2^2,
\end{equation}
which suppresses temporal fluctuations caused by noise or environmental interference.

\subsubsection{Classifier-level Representation Smoothing}  
In addition to feature alignment, we perform optimization at the classifier level to mitigate decision boundary instability under domain shift. Specifically, given the classifier output logits $\mathbf{p}_{\text{CSI}}$ for WiFi input and $\mathbf{p}_{\text{LM}}$ for the semantic teacher, we design a smoothed KL divergence objective:
\begin{equation}
\mathcal{L}_{\text{cls}} = \text{KL}\big(\sigma(\mathbf{p}_{\text{CSI}}/\tau) \;\|\; \sigma(\mathbf{p}_{\text{LM}}/\tau)\big)
\end{equation}
where $\sigma(\cdot)$ denotes the softmax operator and $\tau$ is a temperature parameter. This loss encourages WiFi predictions to remain consistent with semantically structured teacher outputs, reducing overconfidence on ambiguous gesture classes.

\subsubsection{Overall Objective of MARO}  
The final optimization objective for MARO combines feature-level alignment, temporal consistency, and classifier smoothing:
\begin{equation}
\mathcal{L}_{\text{MARO}} = \lambda_1 \mathcal{L}_{\text{feat}} + \lambda_2 \mathcal{L}_{\text{temp}} + \lambda_3 \mathcal{L}_{\text{cls}},
\end{equation}
where $\lambda_1, \lambda_2, \lambda_3$ are balancing hyperparameters. This dual-view alignment strategy jointly reduces cross-modal discrepancy and domain-induced variations, enabling robust and discriminative WiFi gesture recognition in unseen environments.

\subsection{Training Objective and Optimization}

The overall objective of \name~is to integrate semantic supervision from large models with modality alignment strategies in order to enhance both representation generalization and deployment efficiency. Specifically, the training loss consists of two main components, LSDM and MARO.

LSDM enforces semantic alignment between WiFi embeddings $\mathbf{z}_{\text{CSI}}$ and teacher embeddings $\mathbf{z}_{\text{LM}}$ via a contrastive distillation loss:
$\mathcal{L}_{\text{LSDM}}.$
MARO further reduces distributional discrepancy and temporal inconsistency through three complementary terms. First, the feature-level alignment loss $\mathcal{L}_{\text{feat}}$ encourages the CSI embeddings to remain close to the distribution centers of teacher embeddings, thereby reducing modality-induced drift. Second, the temporal consistency loss $\mathcal{L}_{\text{temp}}$ constrains consecutive CSI segments to be smooth in feature space, suppressing noise-induced fluctuations and improving stability. Finally, the classifier smoothing loss $\mathcal{L}_{\text{cls}}$ regularizes the prediction outputs of the WiFi encoder by aligning them with the semantically structured teacher distributions, which mitigates overconfidence and enhances discriminability under ambiguous gesture categories.
The MARO objective is formulated as:
\begin{equation}
\mathcal{L}_{\text{MARO}} = \lambda_1 \mathcal{L}_{\text{feat}} + \lambda_2 \mathcal{L}_{\text{temp}} + \lambda_3 \mathcal{L}_{\text{cls}},
\end{equation}

where $\lambda_1, \lambda_2, \lambda_3$ are balancing coefficients.
The final training objective combines both modules:
\begin{equation}
\mathcal{L}_{\text{total}} = \mathcal{L}_{\text{LSDM}} + \mathcal{L}_{\text{MARO}}.
\end{equation}

This unified objective allows the WiFi encoder to simultaneously inherit semantic priors from pre-trained foundation models and adapt to domain-invariant alignment constraints.

We adopt mini-batch stochastic gradient descent with momentum to optimize all parameters of the WiFi encoder and classifier. The large model teacher is kept frozen during training, serving only as a semantic teacher. In practice, we empirically set $\lambda_1, \lambda_2, \lambda_3$ to balance semantic transfer and alignment stability. This optimization pipeline ensures that the student WiFi encoder learns semantically grounded and robust gesture representations suitable for efficient deployment in AIoT environments.

\begin{table*}[htbp]
\renewcommand{\arraystretch}{1.5}
\centering
\small
\setlength{\tabcolsep}{5mm}
\caption{Comparison (\%) of accrucy before and after knowledge distillation}
\label{tab:distillation-ablation-en}
\begin{tabular}{llcccc}
\toprule
\textbf{Feature} & \textbf{Stage} & \textbf{ID} & \textbf{CL} & \textbf{CO} & \textbf{Mean} \\
\midrule
\multirow{2}{*}{\textbf{DFS}} & baseline & 88.16 & 82.00 & 78.83 & 83.00 \\
& our method & \textbf{90.61} \textcolor{teal}{(+2.45)} & \textbf{83.78} \textcolor{teal}{(+1.78)} & \textbf{81.33} \textcolor{teal}{(+2.50)} & \textbf{85.24} \textcolor{teal}{(+2.24)} \\
\midrule
\multirow{2}{*}{\textbf{CSI-Ratio}} & baseline & 95.56 & 94.28 & 89.92 & 93.25 \\
& our method & \textbf{97.78} \textcolor{teal}{(+2.22)} & \textbf{95.59} \textcolor{teal}{(+1.31)} & \textbf{92.80} \textcolor{teal}{(+2.88)} & \textbf{95.39} \textcolor{teal}{(+2.14)} \\
\bottomrule
\end{tabular}
\end{table*}

\section{EXPERIMENTS}
\subsection{Experimental Setup and Implementation Details}
\subsubsection{Dataset and Preprocessing}We validate our method on the public Widar3.0 dataset \cite{zhang2018widar3}, a large-scale CSI-based gesture dataset. To simulate realistic single-receiver conditions, we exclusively use data collected from one receiver location. We conduct both intra-domain and cross-domain (cross-location and cross-orientation) experiments. For intra-domain evaluation, data is split into training and validation sets with a 4:1 ratio. For cross-domain tests, the model is trained on data from one location or direction and tested on all others. Prior to training, the raw CSI data undergoes a multi-stage preprocessing pipeline, including Hampel filtering for outlier removal, phase unwrapping to correct ambiguities, and resampling to a uniform sequence length to normalize gesture duration. Specifically, this pipeline is designed to extract two complementary feature modalities from the signal. The gestural component is then represented by: 1) the Doppler Frequency Shift (DFS), which captures motion-induced spectral changes, and 2) the CSI-Ratio, which encodes phase variations across the antenna array.

\subsubsection{Evaluation Metric and Baselines} We adopt classification accuracy as the primary evaluation metric. To demonstrate the effectiveness of our proposed framework, we benchmark our method against several state-of-the-art (SOTA) competitors. Additionally, we conduct a detailed ablation study, comparing the performance of our student model before and after knowledge distillation, to specifically quantify the improvements imparted by our strategy.

\subsubsection{Implementation Details}
All experiments are implemented using the PyTorch framework and conducted on a single NVIDIA RTX 3090 GPU. The model is trained for 100 epoches with a batch size of 16 using the Adam optimizer. The initial learning rate is set to ${4}\times{10^{-5}}$ and adjusted using a cosine scheduler. For knowledge distillation, we employ the pre-trained GPT-2 model as the teacher model. The student model's total training loss is a well-designed composite objective function. It is formulated as a weighted sum of four distinct components: the supervised cross-entropy losses for the teacher and student branches, the intermediate feature alignment loss, and the soft-label distillation loss.

\subsection{Training and Inference Strategy}
Our training and inference strategies are modality-specific to optimally leverage the distinct characteristics of CSI-Ratio phase and DFS data. For CSI-Ratio data, we treat each antenna’s stream as an independent sample during training. Then, at the inference process, the gesture probability distributions from each antenna-specific sample are averaged to form a fused distribution. The final prediction is the class with the highest confidence in this fused result. For DFS data, the entire feature set is treated as a single, holistic sample for the training process. While at the inference process, the model's prediction is determined directly by the class with the maximum confidence in the output probability distribution.

\subsection{Comparison with baseline and the State-of-the-art Methods}
 
We conduct a series of experiments to demonstrate the performance gains from our proposed knowledge distillation framework by comparing the classification accuracy of the small model against its own baseline version prior to distillation.
The experimental results are shown in Tab.~\ref{tab:widar-results}. It can be seen that our method achieves the best results in accuracy. As shown in Tab.~\ref{tab:distillation-ablation-en}, our proposed method achieves a 1.31\%-2.88\% improvement in accuracy on the Widar 3.0 dataset. This demonstrates that the teacher model provides a consistent performance boost to the student model within our framework, regardless of the underlying feature data.

\begin{figure*}[htbp]
    \centering
    
    \begin{minipage}{1.0\textwidth}
        
        \begin{subfigure}{0.32\linewidth}
            \centering
            \includegraphics[width=\linewidth]{"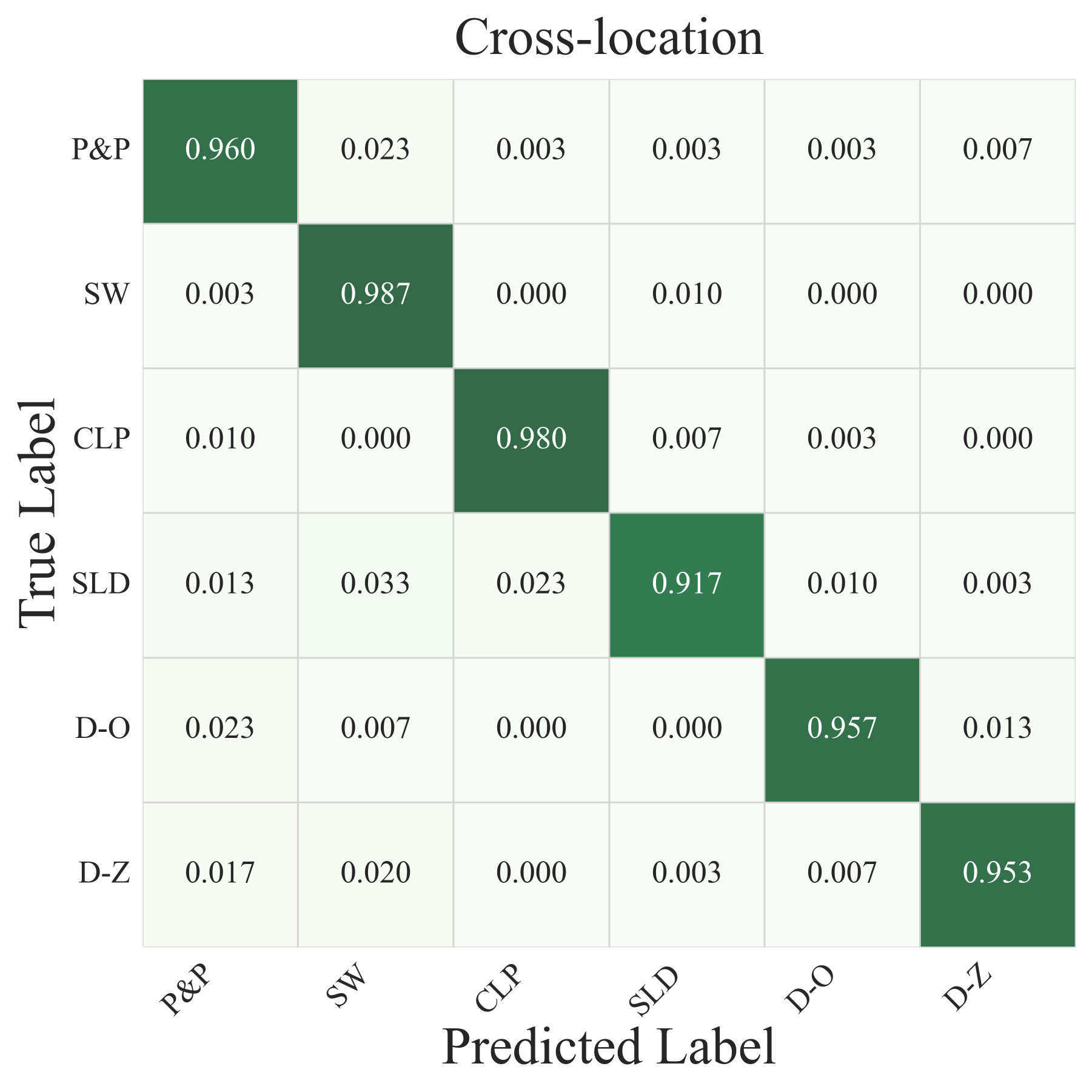"}
            \caption{Cross-location}
            \label{fig:sub1}
        \end{subfigure}
        \hfill 
        \begin{subfigure}{0.32\linewidth}
            \centering
            \includegraphics[width=\linewidth]{"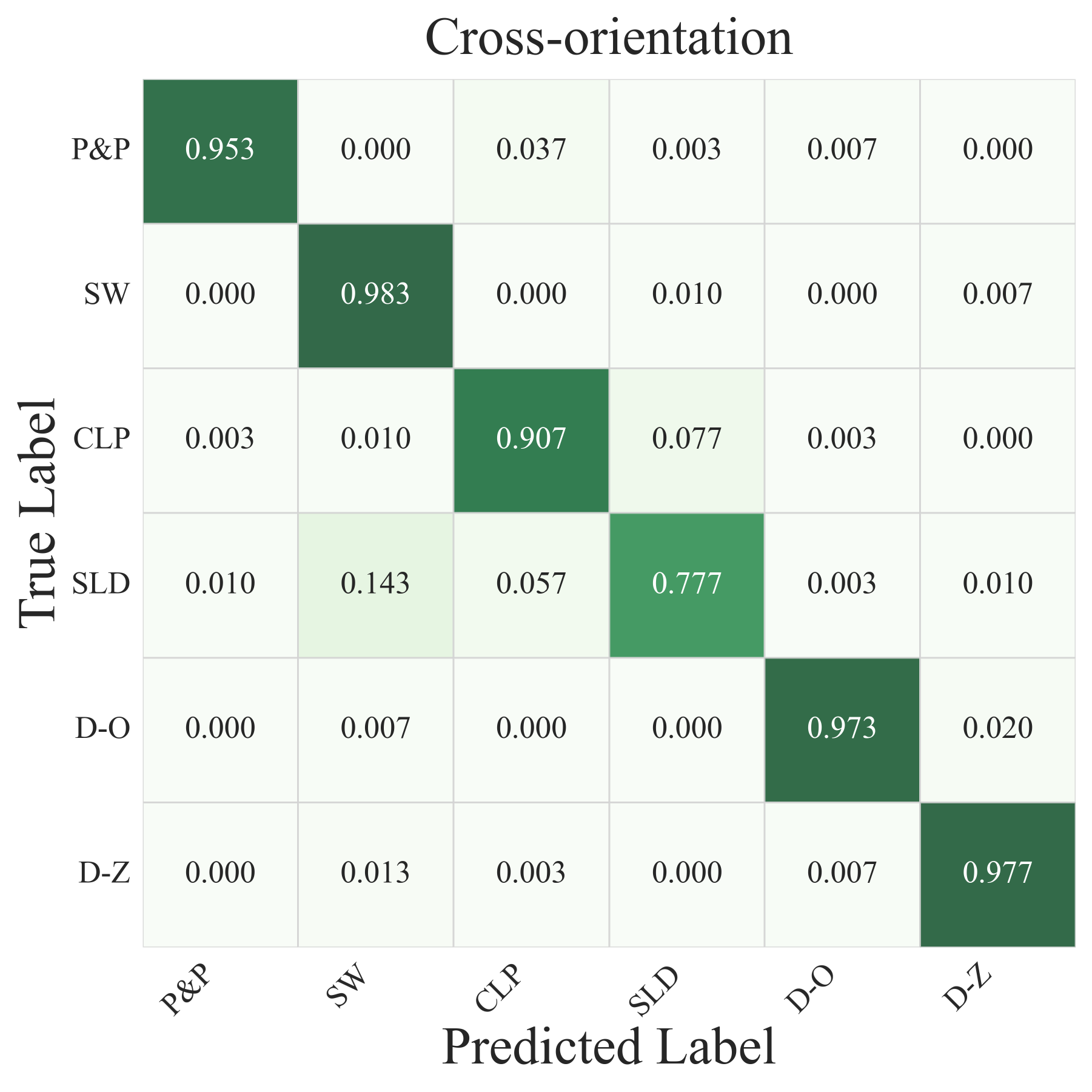"}
            \caption{Cross-orientation}
            \label{fig:sub2}
        \end{subfigure}
        \hfill 
        \begin{subfigure}{0.32\linewidth}
            \centering
            \includegraphics[width=\linewidth]{"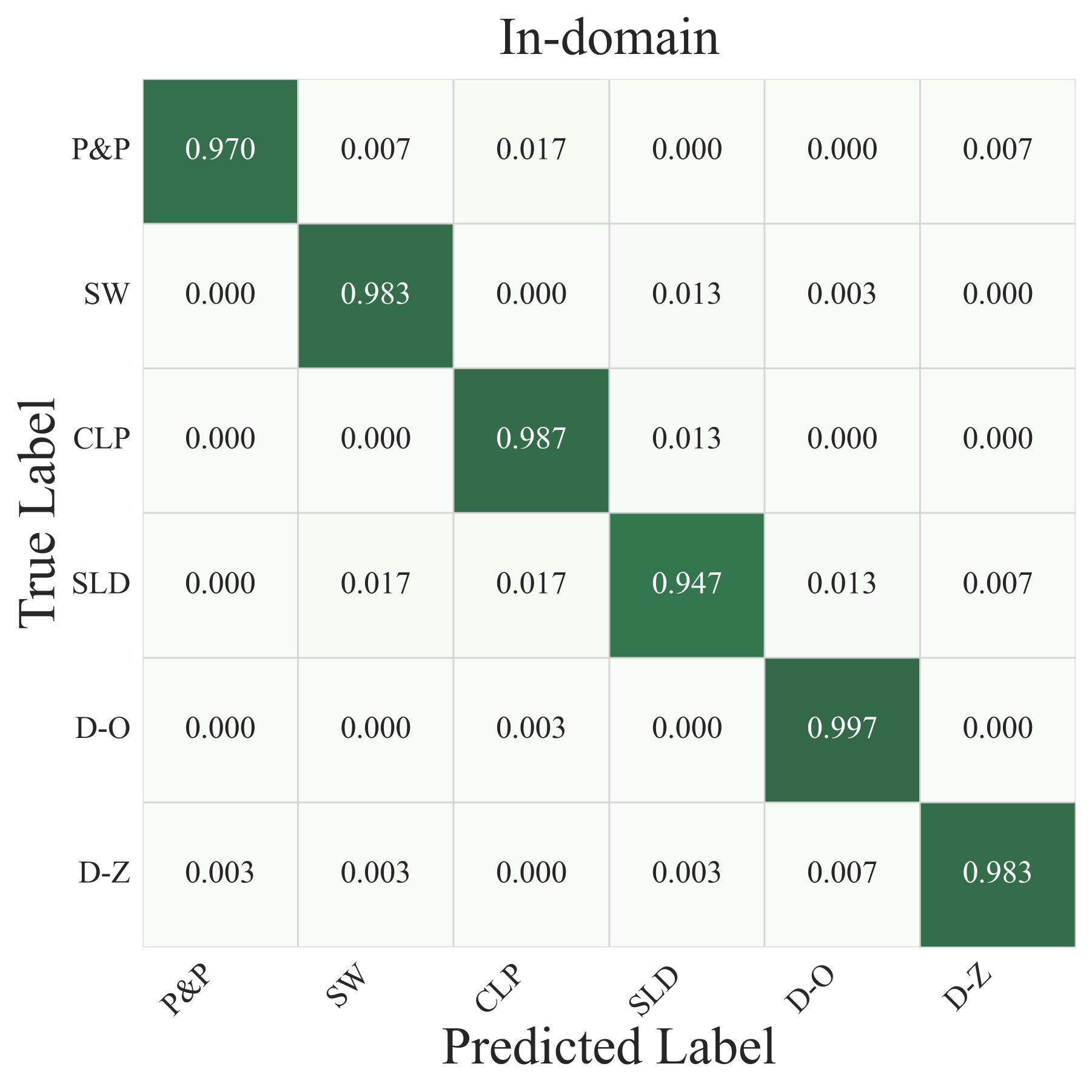"}
            \caption{In-domain}
            \label{fig:sub3}
        \end{subfigure}
    \end{minipage}

    \caption{\centering{Confusion Matrices Under Different Scenarios, where P\&P, SW, CLP, SLD, D-O, and D-Z correspond to the gestures Push\&Pull, Sweep, Clap, Slide, Draw-O, and Draw-ZigZag, respectively.}}
    \label{fig:main}
\end{figure*}

\subsection{Ablation Studies and Visualization}
To validate the efficacy of our proposed components, we conduct a series of ablation studies on the Widar 3.0 dataset. We aim to answer two primary questions: (1) How effective is our framework in enhancing the student model's generalization capability? (2) Which data preprocessing method, CSI-Ratio Phase or DFS, yields a more robust feature representation for gesture recognition tasks? 

\subsubsection{Effect of Knowledge Distillation}

To isolate the contribution of our proposed cross-modal knowledge distillation framework, we compare the performance of our full model against a baseline. The baseline model consists of the same student network architecture but is trained independently without the guidance of the large language model teacher and the distillation losses. The results are presented in Tab.~\ref{tab:distillation-ablation-en}. The student model, when guided by our framework, demonstrates significant performance gains across all evaluation settings. For the CSI-Ratio Phase preprocessing method, the distilled model achieves accuracy improvements of 2.22\%, 1.31\%, and 2.88\% in the intra-domain, cross-position, and cross-direction scenarios, respectively. A similar trend is observed for the DFS preprocessing method, where the model's accuracy increases by 2.45\% (intra-domain), 1.62\% (cross-position), and 2.50\% (cross-direction) after distillation.

These results strongly indicate that transferring knowledge from the pre-trained large model via our dual distillation strategy effectively enhances the student model's generalization ability. The improvement is particularly noticeable in the more challenging cross-domain scenarios, which directly confirms that our framework has a great capacity to helps the student model learn more robust and domain-invariant features.

\subsubsection{Impact of Preprocessing Methods}

As shown in Tab.~\ref{tab:distillation-ablation-en}, the choice of data preprocessing significantly impacts model performance. Across all experiments, both with and without distillation, the CSI-Ratio Phase method consistently outperforms the DFS method. For instance, our final model using CSI-Ratio Phase achieves a cross-direction accuracy of 92.80\%, which is substantially higher than the 81.33\% achieved with DFS.

This suggests that the CSI-Ratio Phase representation preserves more discriminative information from the raw CSI signals and exhibits greater robustness to environmental variations. The phase-based features appear to be more sensitive to the nuanced changes induced by hand gestures, making them a more suitable input for high-accuracy recognition tasks.

\subsection{Experimental Analysis and Discussion}

To comprehensively evaluate the performance of our proposed framework, we conduct both quantitative and qualitative analyses across three distinct scenarios: in-domain, cross-location, and cross-orientation. The empirical results, summarized in Tab.~\ref{tab:distillation-ablation-en}, provide compelling quantitative evidence of the efficacy of our knowledge distillation approach. Our evaluation involves six gesture classes: `Push\&Pull`, `Sweep`, `Clap`, `Slide`, `Draw-O`, and `Draw-ZigZag`, the confusion matrices with in our framework is shown as Fig.~\ref{fig:main}.

\begin{figure}[htbp]
    \centering
    \includegraphics[width=1\linewidth]{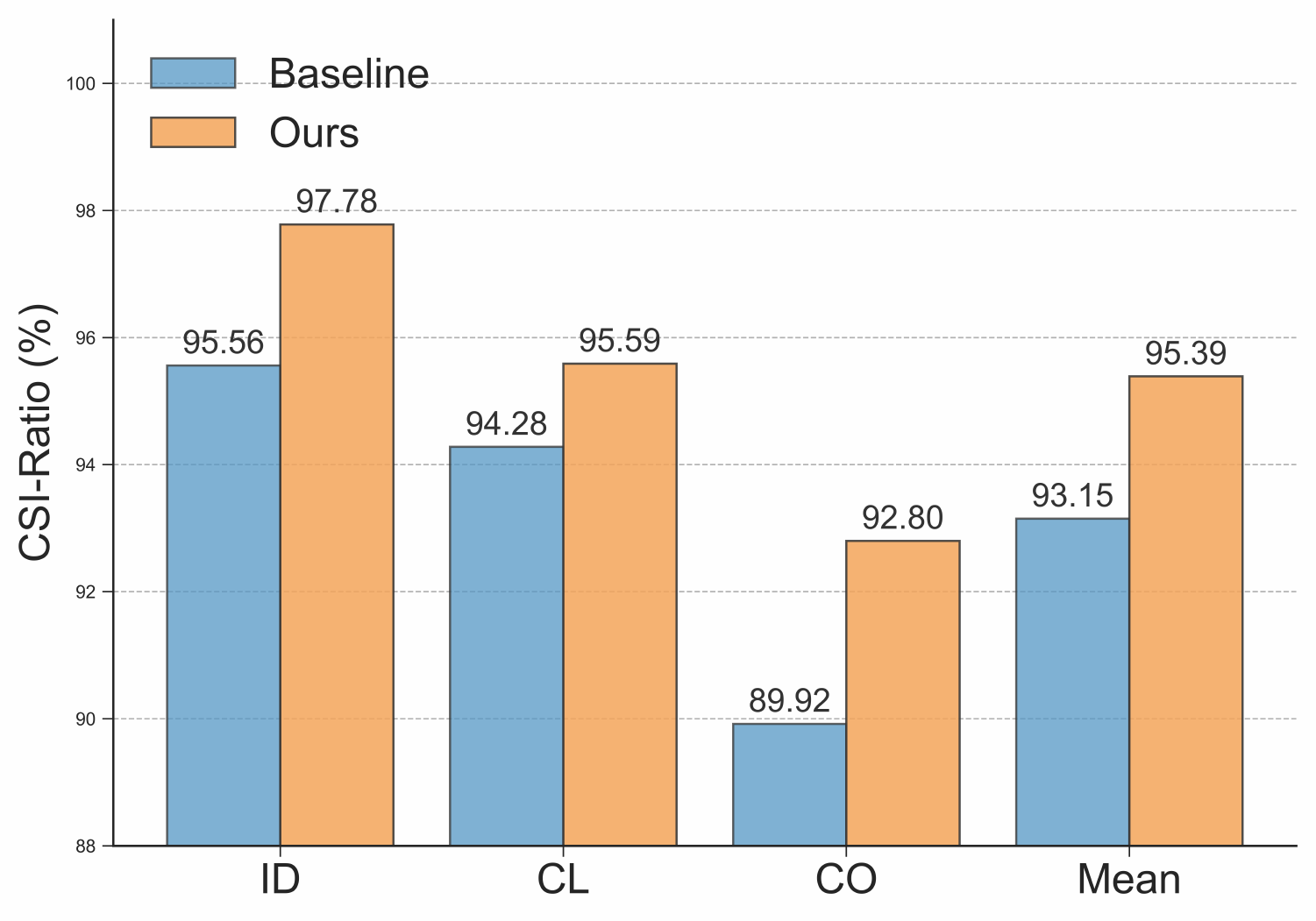}
    \caption{Comparison: Baseline vs. Our Method}
    \label{fig:placeholder}
\end{figure}

As quantitatively demonstrated in Tab.~\ref{tab:distillation-ablation-en}, our proposed framework significantly enhances the model's accuracy across all evaluation scenarios, which can be seen as Fig.~\ref{fig:placeholder}. Focusing on the more robust CSI-Ratio feature set, our method elevates the mean accuracy from 93.25\% to \textbf{95.39\%}, achieving a notable improvement of \textbf{2.14\%}. The performance gain is consistent across all test conditions, with accuracy increasing by 2.22\% in the in-domain setting and 1.31\% in the cross-location setting. This further confirms the model's strong capability to learn discriminative features.

The efficacy of our approach is particularly pronounced in addressing the critical challenge of cross-domain generalization. For the most challenging \textbf{cross-orientation} scenario, where traditional models often falter, knowledge distillation yields the most substantial accuracy gains: an impressive \textbf{+2.88\%} for CSI-Ratio and \textbf{+2.50\%} for DFS. This highlights our framework's strength in improving model robustness against domain shifts caused by user orientation changes.

Traditional approaches in this domain often rely on compact models trained from scratch, which typically exhibit weaker feature learning capabilities. This limitation is empirically validated by the baseline model's performance, which stagnates at 89.92\% (CSI-Ratio) in the cross-orientation task. In stark contrast, our distillation framework pushes this boundary to \textbf{92.80\%}. This significant improvement stems from the core mechanism of our approach: the pre-trained large teacher model, with its vast parameter space, provides a rich and generalized feature manifold. Through our distillation strategy, the student model is guided to emulate the teacher's sophisticated feature representations, endowing it with enhanced discriminative power. This enables it to capture the fine-grained differences between kinematically similar gestures, even under significant domain shifts. Therefore, the quantitative results in Tab.~\ref{tab:distillation-ablation-en} robustly validate our central hypothesis: knowledge distillation from large models is a highly effective paradigm for enhancing the generalization and robustness of specialized models in Wi-Fi-based sensing tasks.

\section{Conclusion}
We propose a novel framework, \name, to enhance WiFi-based gesture recognition performance by addressing the challenges of domain sensitivity, limited semantic abstraction, and lack of robust supervision. By integrating large-model semantic distillation with modality-aligned representation optimization, our framework overcomes the limitations of existing methods, which often struggle with poor cross-domain generalization and ambiguous gesture classification. Specifically, we introduce the Large-model Semantic Distillation Module (LSDM), which transfers high-level conceptual knowledge from pre-trained large models to the WiFi domain, thereby improving the semantic grounding of CSI representations. To further reduce modality discrepancy and enhance robustness, we design the Modality-Aligned Representation Optimization Module (MARO), which jointly enforces feature alignment, temporal consistency, and classifier-level smoothing. These components work synergistically to build a generalizable and lightweight WiFi recognition pipeline. In future work, we plan to explore more advanced semantic supervision strategies, incorporate multimodal pretraining paradigms, and extend our framework to broader RF sensing tasks in AIoT applications.

\section*{ACKNOWLEDGEMENTS}
This work is supported by Anhui Province Science Foundation for Youths (Grant No. 2308085QF230), Fundamental Research Funds for the Central Universities (Grant No. JZ2025HGTB0225), Major Scientific and Technological Project of Anhui Provincial Science and Technology Innovation Platform (Grant No. 202305a12020012), National Natural Science Foundation of China (Grant No. 62302145). 
\bibliographystyle{IEEEbib}
\bibliography{icme2025references}


\end{document}